\documentclass[10pt,twocolumn,letterpaper]{article}

\usepackage{cvpr}
\usepackage{times}
\usepackage{epsfig}
\usepackage{graphicx}
\usepackage{amsmath}
\usepackage{amssymb}

\usepackage{booktabs}

\usepackage[pagebackref=true,breaklinks=true,letterpaper=true,colorlinks,bookmarks=false]{hyperref}

\cvprfinalcopy 

\def\cvprPaperID{3101} 

\ifcvprfinal\fi
\begin{document}

\title{Focusing and Diffusion: Bidirectional Attentive Graph Convolutional Networks \\for Skeleton-based Action Recognition}

\author{Jialin Gao$^{1}$,\ \ Tong He$^{2}$,\ \ Xi Zhou$^{1}$, \ \ Shiming Ge$^3$\\
$^{1}$Shanghai Jiao Tong University\ \ \ \ $^{2}$University of California, Los Angeles\\
$^3$Institute of Information Engineering, Chinese Academy of Sciences \\
{\tt \small jialin\_gao@sjtu.edu.cn, simpleig@cs.ucla.edu, zhouxi@cloudwalk.cn, geshiming@iie.ac.cn}
}


\maketitle

\begin{abstract}
	A collection of approaches based on graph convolutional networks have proven success in skeleton-based action recognition by exploring neighborhood information and dense dependencies between intra-frame joints. However, these approaches usually ignore the spatial-temporal global context as well as the local relation between inter-frame and intra-frame. In this paper, we propose a focusing and diffusion mechanism to enhance graph convolutional networks by paying attention to the kinematic dependence of articulated human pose in a frame and their implicit dependencies over frames. In the focusing process, we introduce an attention module to learn a latent node over the intra-frame joints to convey spatial contextual information. In this way, the sparse connections between joints in a frame can be well captured, while the global context over the entire sequence is further captured by these hidden nodes with a bidirectional LSTM. In the diffusing process, the learned spatial-temporal contextual information is passed back to the spatial joints, leading to a bidirectional attentive graph convolutional network (BAGCN) that can facilitate skeleton-based action recognition. Extensive experiments on the challenging NTU RGB+D and Skeleton-Kinetics benchmarks demonstrate the efficacy of our approach.
\end{abstract}

\section{Introduction}

\begin{figure}[t]
	\begin{center}
		\includegraphics[width=1.0\linewidth]{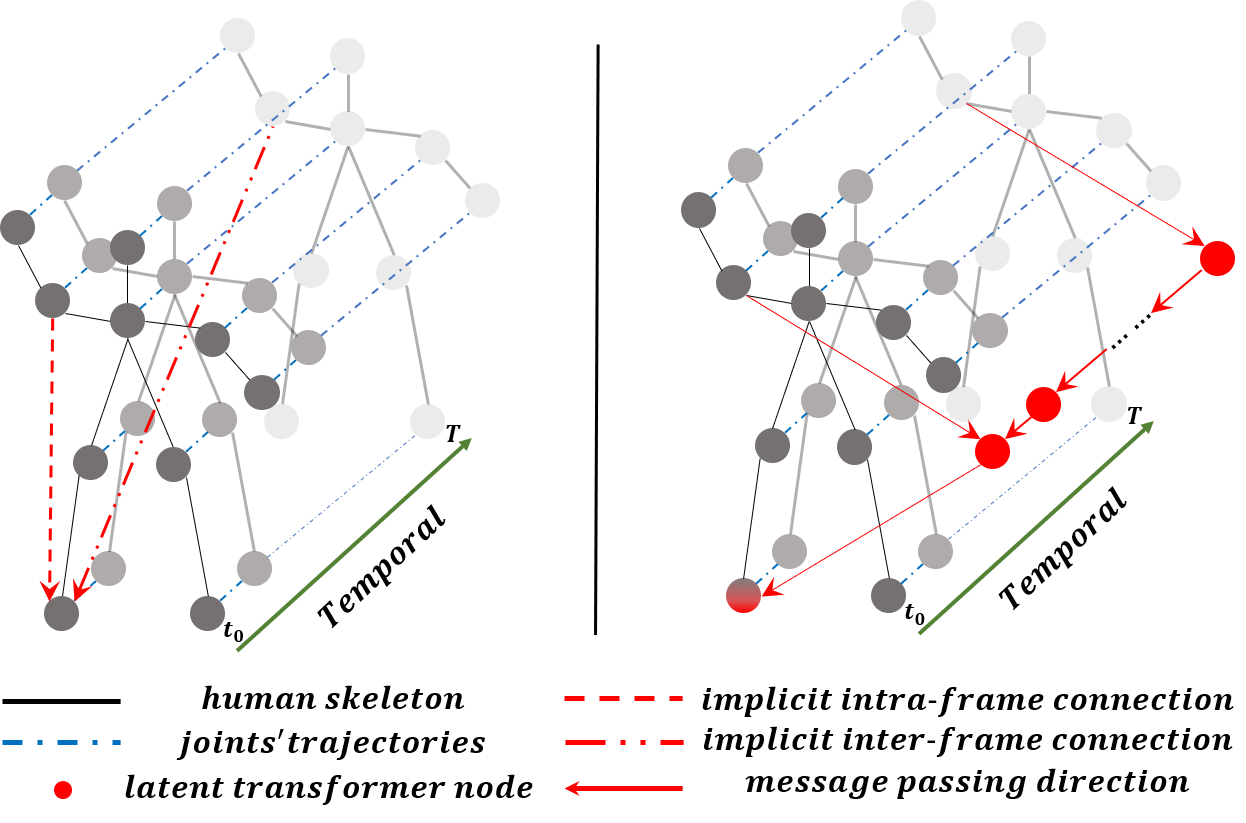}
	\end{center}
	\caption{Previous methods ignore the implicit inter-frame and intra-frame connections, which inspired us to learn a latent transformer node and then convey the global spatial-temporal context information back to spatial joints in each frame.}
	\label{fig:motivation}
\end{figure}
Action recognition is a challenging vision problem that has been studied for years. It plays an essential role in various applications, such as video surveillance \cite{mehran2009abnormal}, patient monitoring \cite{liu2015simple}, robotics\cite{yun2012two}, human-machine interaction\cite{jiang2015human} and so on~\cite{wang2016temporal, yun2012two,shahroudy2016ntu}. Previous methods mainly rely on two different data modalities: RGB images \cite{kay2017kinetics, soomro2012ucf101} and skeleton points \cite{shahroudy2016ntu, yan2018spatial,yun2012two}. Compared with 2D images, 3D (skeleton) points directly encode object shape and motion~\cite{qi2017pointnet,qi2017pointnet++,wu2019comprehensive,he2019geonet,liu2019flownet3d,torresani2004learning,He2019Mono3DM3} and thus are more robust to nuisances like fast motions, camera viewpoint changes, etc.

Conventional attempts \cite{vemulapalli2014human, fernando2015modeling, liu2017enhanced} take skeletons as a sequence of vectors generated by encoding all joints in the same frame together. Some deep-learning-based methods, such as CNNs, employ convolution operation directly on skeletons, the irregular grid data. These approaches ignore the intrinsically kinematic dependencies and structural information of human pose, leading to unsatisfactory performance.

To exploit the structure of skeleton data, Yan \etal \cite{yan2018spatial} first introduces the spatial-temporal Graph Convolutional Network (GCN) based on coordinated joints and the natural topology of the human body. However, this predefined graph structure suffers from two defects, as shown in Fig.\ref{fig:motivation}: 1) the graph only considers physically structural information but ignores the semantic connections among intra-frame joints. For example, the implicit correlation between hands and head is crucial to recognize the action of "drinking." 2) the inter-frame edges only connect the same joints over consecutive frames, which lack modeling the connections between inter-frame joints. Another challenge is that a complete action lasts from several seconds to minutes. Thus, it is crucial to design a model that is capable of reducing noise from irrelevant joints and focusing on informative ones. Temporal context modeling is usually expected to be helpful for addressing this issue, inspired by which in this paper we intend to capture both the physical connections and semantic dependencies beyond the spatial and temporal neighborhood restrictions of GCN. 


Several methods, like AS-GCN\cite{li2019actional} and 2s-AGCN\cite{shi2019two}, attempt to overcome the aforementioned first drawback but they introduce additional challenges to be solved. For obtaining kinematic dependency, 2s-AGCN introduces a module to learn the additional data-driven morphing graph over layers, and AS-GCN proposes an encoder-decoder structure to capture action-specific dependencies combined with physically structural links between joints. Nevertheless, it is inefficient to directly model the relationship between non-neighbor joints over a fully-connected graph. These potentially dense connections are also tricky to learn because of the noise from irrelevant ones, which leads to difficulty in optimization and unsatisfied performance. Besides, they also ignore the correlations between inter-frame joints so that they cannot take advantage of the temporal contextual information.

To overcome such limitations, we introduce a focusing and diffusion mechanism to enhance the graph convolutional networks to receive information beyond the spatial and temporal neighborhood restrictions. We first represent the human skeleton in each frame by constructing a focus graph and diffusion graph. Then we propose an efficient way to propagate information over these graphs though a learnable latent transformer node. This way captures a sparse but informative connections between joints in each frame, since not all the implicit correlations are activated in an action. It also agrees with the fact that informative connections may also change over the frames. During the training stage, the latent transformer node learns how to connect the action-specific related joints and passes message forth and back though the focusing and diffusion graphs.

However, there remains a problem of how to model the relation between intra-frame and inter-frame joints. To solve this, we take advantage of the latent transformer node again. We empirically show that the node learns the informative spatial context in each frame, generating a sequence of latent spatial nodes. To capture the temporal contextual information over frames, we introduce a context-aware module consisting of bidirectional LSTM cells to model temporal dynamics and dependencies based on the learned spatial latent nodes.

With the focusing and diffusion mechanism, a graph convolutional network can selectively focus on the informative joints in each frame, capture the global spatial-temporal contextual information and further convey it back to augment node embedding in each frame. We also develop a Bidirectional Attentive Graph Convolutional Network (BAGCN) as an exemplar and evaluate its effectiveness on two challenging public benchmarks: NTH-RGB+D and Skeleton-Kinetics. In summary, our main contributions are three-fold:

\begin{itemize}
	\item[1)] We propose a novel representation of the human skeleton data in a single frame by constructing two opposite-direction graphs and also introduce an efficient way of message passing in the graph.
	\item[2)] We design a new network architecture, Bidirectional Attentive Graph Convolutional Network (BAGCN), that learns spatial-temporal context from human skeleton sequences leveraging a graph convolutional networks based focusing and diffusion mechanism.
	\item[3)] Our BAGCN model is compared against other state-of-the-arts methods on the challenging NTU-RGB+D and Skeleton-Kinetics benchmarks, and demonstrate superior performance.
\end{itemize}

\section{Related Work}

\textbf{Skeleton-based action recognition:} The methods can be divided into handcraft feature based \cite{wang2012mining, vemulapalli2014human, hussein2013human} and deep learning based \cite{shahroudy2016ntu, kim2017interpretable, li2017skeleton, yan2018spatial}. The former encounters challenges in designing representative features, which results in limited recognition accuracy. But the latter is data-driven and improves the performance by a great margin. There are three types of deep models used widely: RNNs, CNNs and GCNs. RNN-based \cite{du2015hierarchical,shahroudy2016ntu, liu2016spatio, song2017end} methods usually take the skeletons as a sequence of vectors and models their temporal dependencies over frames. CNN-based approaches \cite{kim2017interpretable, ke2017new,liu2017enhanced,li2017skeleton,li2018co} employ convolution on skeleton data by regarding it as a pseudo-image. 

Recently, graph-based methods \cite{shi2019skeleton, shi2019two, yan2018spatial, li2019actional} perform superior classification accuracies. ST-GCN \cite{yan2018spatial} is the first one to employ graph convolution. SR-TSL \cite{si2018skeleton} uses gated recurrent unit (GRU) to propagate messages on graphs and uses LSTM to learn the temporal features. AS-GCN \cite{li2019actional} learns actional links simultaneously with structural links in their graph model. 2s-AGCN\cite{shi2019two} introduces an adaptive module to construct a data-driven morphing graphs over layers. However, these two methods meet the same difficulties in optimization due to noise from irrelevant joints or bones. Further, DGNN\cite{shi2019skeleton} constructs a directed graph and define the message passing rules for node and edge updating. However, they ignore the implicit correlations between intra-frame and inter-frame joints.

\textbf{Graph convolutional networks:} Graph convolutional networks
(GCNs), which generalizes convolution from image to graph, has been successfully applied in many applications. The way of constructing GCNs on graph can be split into 2 categories: the spectral \cite{henaff2015deep, li2015gated, kipf2016semi} and spatial \cite{bruna2013spectral, niepert2016learning} perspective. The former utilizes graph Fourier transform performing convolution in frequency domain. Differently, the latter directly performs the convolution filters on the graph according to predefined rules and neighbors. We follow the second way to construct the CNN filters on the spatial domain.

\section{Method}
To obtain the physically and semantically connections sparsely, we introduce a focusing and diffusion mechanism to enhance the graph convolutional networks' ability in receiving information from other joints of intra-frame and inter-frame. First, we define focusing and diffusion graphs based on the skeletons. Then we formulate how to convey message over constructed graphs. Afterwards, we introduce an exemplar of our focusing and diffusion mechanism, developing a bidirectional attentive graph convolutional networks. 

\begin{figure}[t]
	\begin{center}
		\includegraphics[width=0.9\linewidth]{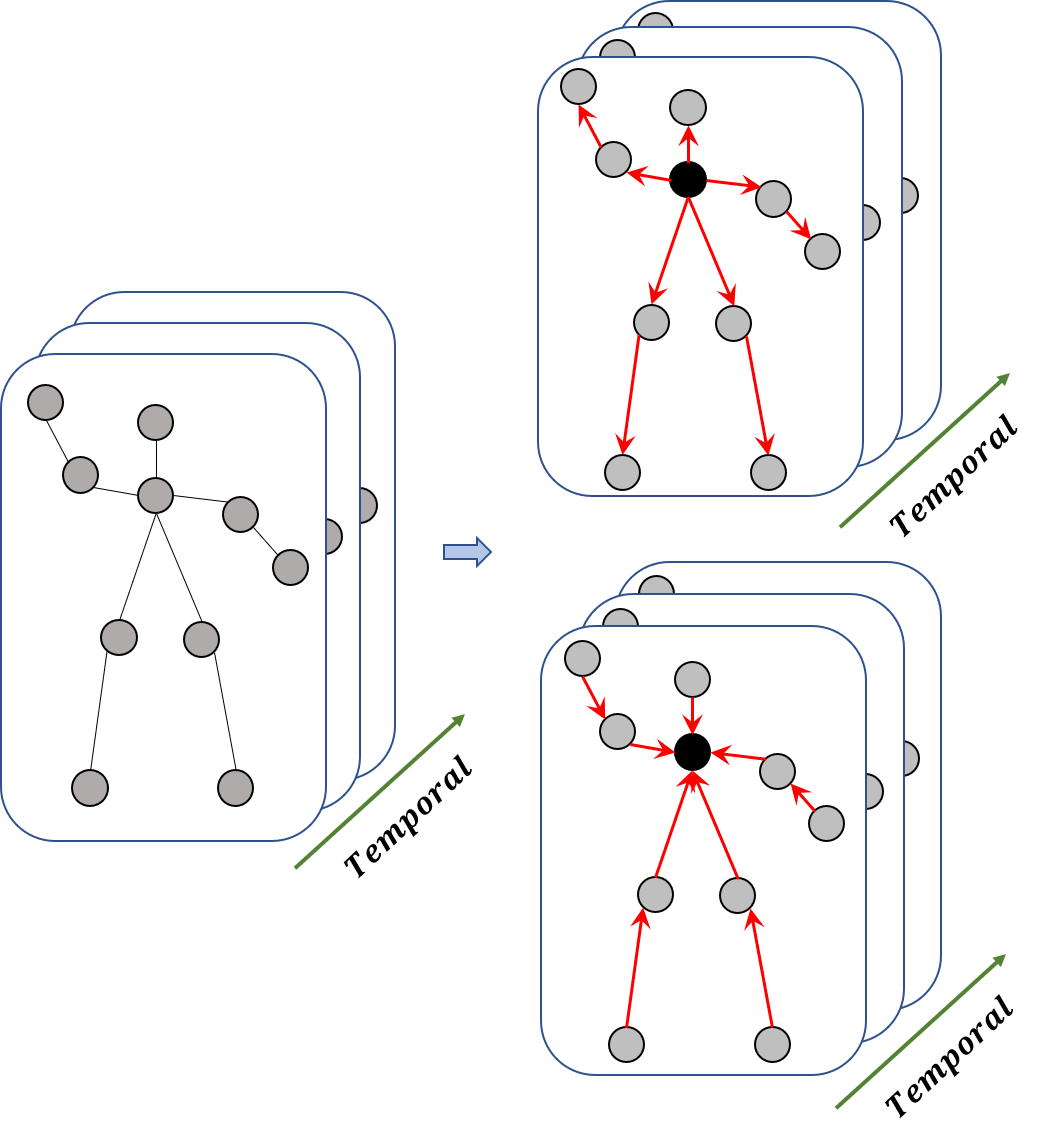}
	\end{center}
	\caption{Undirected graphs (left) can be modeled by explicitly assigning two
		directed edges in opposite direction for each undirected edge, leading to two directed graphs(right). The black node denotes the center node and arrow indicates the passing direction.}
	\label{fig:graph_construct}
\end{figure}
\subsection{Graph Definition}
The raw skeleton data consists of a sequential articulated human pose coordinates. Previous works \cite{li2017skeleton, shi2019two} construct an undirected spatial temporal graph the same as \cite{yan2018spatial}. Given a spatial-temporal graph $\mathcal{G}=\{\mathcal{V},\mathcal{E}\}$, where $\mathcal{V}=\{v^t_{i}|t=1,\cdots,T, i=1,\cdots,V\}$ is the set of $V$ body joints and $\mathcal{E}=\{\mathcal{E_S}, \mathcal{E_T}\}$ is the set of intra-skeleton and inter-frame bones. $\mathcal{E_S}$ contains the natural connections in articulated human pose and $\mathcal{E_T}$ consists of the joints' trajectories between consecutive two frames. As shown in the left part of Fig.\ref{fig:graph_construct}, the vertexes denote body joints and the naturally connections between them represent the bone. 

Different from previous works, we define an undirected spatial temporal graph as two directed graphs with opposite direction for each edges inspired by \cite{kipf2018neural}, where $\mathcal{G}=\{ \{\mathcal{V},\overrightarrow{\mathcal{E}}\},  \{\mathcal{V},\overleftarrow{\mathcal{E}}\} \}$, as shown in the right part of Fig.\ref{fig:graph_construct}. The message passing between joints is bi-directional, which is expected to exploit the kinematics dependencies and help the center joint and peripheral joints receive information from each other.

ST-GCN \cite{yan2018spatial} indicates that a graph with spatial configuration partitioning strategy exploits more location information. We follow this conclusion and split the neighbors into three subsets: 1) the root node itself; 2) the centripetal group; 3) the centrifugal group. That we have $\mathcal{S}$ = \{root, closer, far\}. It is according to the distance from the gravity center of the skeleton. Hence, we define the message passing in focus graphs from a joint $v_i$ close to the center of gravity to a joint $v_j$ far from the center of gravity and that direction is opposite in diffusion graphs. The bone information $e_{\vec{ij}}$ is obtained by calculating the coordinates difference between these two points.

Let $\mathbf{A} \in \{0,1\}^{n\times n}$ be the adjacency matrix of our graph $\mathcal{G}$, where $\mathbf{A} = \{\mathbf{A^{f}}, \mathbf{A^{d}}\}$ represents the corresponding adjacency matrices to focus graph $\{\mathcal{V},\overrightarrow{\mathcal{E}}\}$ and diffusion graph $\{\mathcal{V},\overleftarrow{\mathcal{E}}\}$ , respectively. For each sub-graph, we have $\sum_{s\in \mathcal{S}} \mathbf{A_s^f}=\mathbf{A^f}$ and $\sum_{s\in \mathcal{S}} \mathbf{A_s^d}=\mathbf{A^d}$

\begin{figure*}[th]
    \vspace{-14mm}
	\begin{center}
		\includegraphics[width=0.9\linewidth]{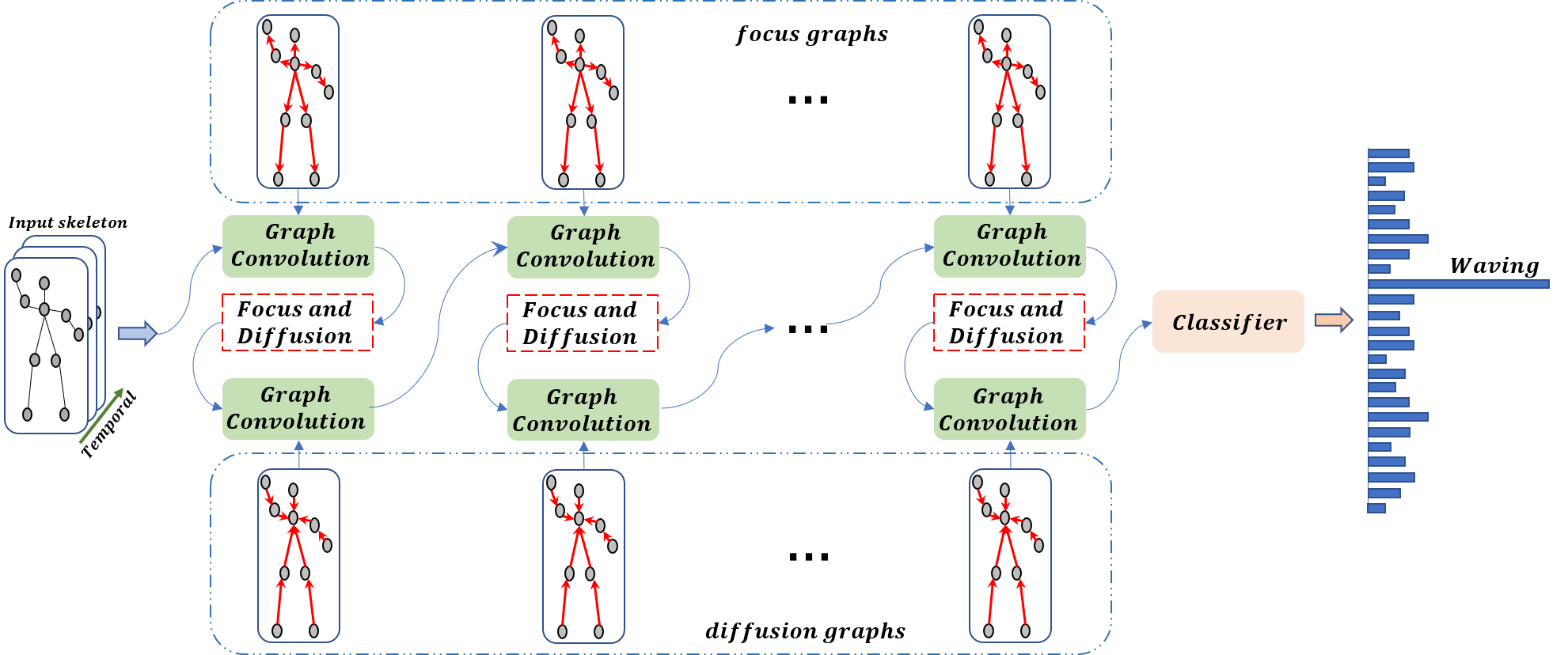}
	\end{center}
	\caption{Illustration of our bidirectional attentive graph convolutional network for skeleton-based action recognition. Besides the stacked spatial-temporal graph convolution, we introduce a focusing and diffusion mechanism to learn a latent transformer node and bidirectionally convey the local context over frames. Once capturing the global spatial-temporal contextual information, spatial joints learn how much context should be absorbed for updating node embeddings in each frame.}
	\label{fig:framework}
\end{figure*}
\subsection{Building Block with Graph Convolution}
We have defined the sequential skeletons as two directed spatial-temporal graphs. The problem becomes how to passing messages in these directed graphs. We adopt a similar implementation of spatial graph convolution as in \cite{kipf2016semi}, which employs a weighted average of neighboring features for updating intra-frame joints' embedding.  According to the definition in $\mathcal{G}$, each graph convolution layer is followed by a temporal convolution, so that we only need to implement graph convolution operation in a single frame case.

For message passing, we first convey information though focus graphs and then transform the updated features back via diffusion graphs. Let $\mathbf{X}_{in} \in \mathbb{R}^{V\times T \times C }$ be the nodes' embedding of our focus graph from a certain intermediate layer. Before the focus stage, the node embedding is first updated via a graph convolution
\begin{equation}
\mathbf{X}_{mid} = \sum_{s\in \mathcal{S}} \mathbf{M_s} \otimes \hat{\mathbf{A}}_s^f \mathbf{X}_{in} \mathbf{W_s},
\end{equation}
where $\hat{\mathbf{A_s^f}} = \mathbf{\Lambda}_s^{\frac{1}{2}} \mathbf{A_s^f} \mathbf{\Lambda}_s^{\frac{1}{2}} \in \mathbb{R}^{V\times V}$ is the normalized adjacent matrix for each group, $\mathbf{M_s} \in \mathbb{R}^{V\times V}$ is a learnable edge importance weighting and $\otimes$ denotes element-wise production between two matrix. $\mathbf{W_s} \in \mathbb{R}^{C\times C' \times 1 \times 1}$ is the weight matrix, which is implemented via $1\times 1$ convolution operation.

Then, these feature maps will be transformed to learn a latent node during the focus stage, which performs a transformer to pass message to arbitrary nodes beyond the temporal and spatial restrictions. This latent node also can be regarded as the spatial context over a single frame graph. In order to capture the temporal contextual information, we propose to learn the context though a context-aware module, denoted as $CAM$. Afterward, the learned spatial-temporal contextual information can be conveyed back to the spatial node in each frame. Every node learns how much context should be used for embedding during the diffusion stage. For notation, $F_f$ and $F_d$ represent the focusing and diffusion function, respectively. Hence, the latent node $G_S$ and temporal context $G_T$ during the focusing and diffusion can be formulated as:
\begin{equation}
\left\{
\begin{aligned}
G_S &= F_f(X_{mid}) \\
G_{ST} &= CAM(G_S) \\
X_g &= F_d(G_{ST}) \\
\hat{X}_{mid} &= F([Xmid, X_g]), \\
\end{aligned}
\right.
\end{equation}
where $F$ indicates a simple fully-connected layer and $[\cdot,\cdot]$ denotes concatenating operation.

The latent node models global spatial-temporal contextual information which is selectively transferred to update the node features in a single frame graph. Given $\hat{\mathbf{X}}_{mid}$ enhanced with our focusing and diffusion mechanism, another graph convolution operates on the diffusion graph $\mathbf{A^{d}}$ so that the updated node embedding as:
\begin{equation}
\mathbf{X}_{out} = \sum_{s\in \mathcal{S}} \mathbf{M_s^{'}} \otimes \hat{\mathbf{A}}_s^d \hat{\mathbf{X}}_{mid} \mathbf{W_s^{'}},
\end{equation}
where $\hat{\mathbf{A_s^d}} = \mathbf{\Lambda}_s^{\frac{1}{2}} \mathbf{A_s^d} \mathbf{\Lambda}_s^{\frac{1}{2}} \in \mathbb{R}^{V\times V}$ is the normalized adjacent matrix for each group in diffusion graphs, $\mathbf{M'_s} \in \mathbb{R}^{V\times V}$ is a learnable edge importance weighting and $\otimes$ denotes element-wise multiplication operator. $\mathbf{W'_s} \in \mathbb{R}^{C'\times C'' \times 1 \times 1}$ is the weight matrix. Similar to the conventional building block, we add a BN layer and a ReLU layer atfer each graph convolution. With this building blocks, we can construct our bidirectional attentive graph convolutional network as shown in Fig.\ref{fig:framework}, the details of which implementation is shown in next session.

%
%

\subsection{Context-aware Graph Focusing and Diffusion}
\begin{figure}[t]
	\begin{center}
		\includegraphics[width=1.0\linewidth]{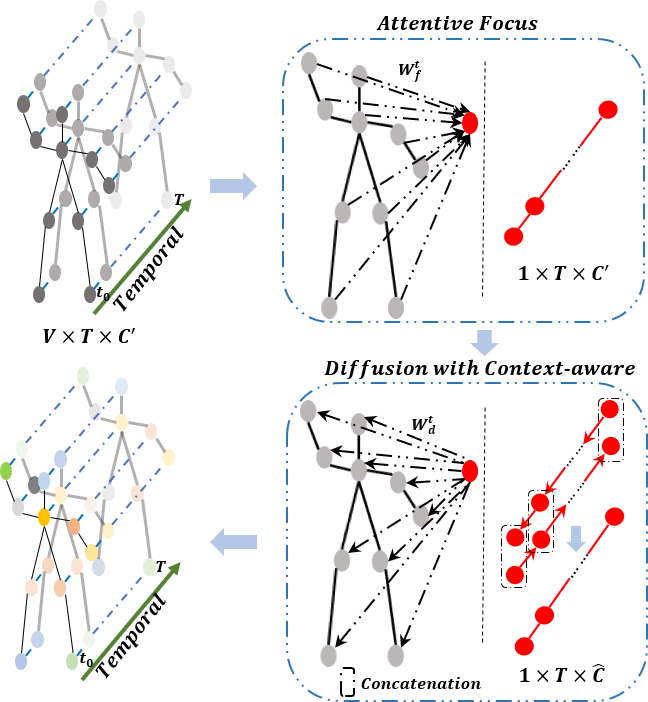}
	\end{center}
	\caption{Implementation details of the context-aware focusing and diffusion.}
	\label{fig:fd}
\end{figure}

In this section, we illustrate an exemplar of our focusing and diffusion mechanism in details, which is shown in Fig.\ref{fig:fd}. First, we introduce attentive focus module and then a temporal context-aware module. After learning the spatial-temporal contextual information with a latent node, we develop a dynamic diffusion to convey the global context back to spatial nodes in each frame.
\subsubsection{Dynamic Attentive Focusing}
The purpose of the focusing stage is to learn a latent node as a transformer, which carries information between nodes in a single frame graph beyond spatial neighborhood  restrictions. Besides, this latent transformer node endows a model with the ability of selectively focusing on the informative joints in each frame. 

Specifically, given an input $\mathbf{f}_{in} \in \mathbb{R}^{V\times T\times C'}$, our attentive focus learns a latent node $G_S\in \mathbb{R}^{1\times T\times C'}$, which satisfies
\begin{equation}
\mathbf{G}^t_S = \mathbf{W}_{f}^t \mathbf{f}_{in}^t \mathbf{W}_{1}
\end{equation}
where $\mathbf{W}_{1} \in \mathbb{R}^{C'\times C'\times 1 \times 1}$ is a $1\times 1$ convolution operation for node embedding and $\mathbf{W}_f \in \mathbb{R}^{1\times T\times V}$ is a learnable matrix to focus the information over a single frame graph. It is equivalent to the attention mechanism combining features from all nodes over a spatial graph. In our spatial-temporal graphs, it is also temporally dynamic and each spatial graph in a sequence has its own attention causing a series of global spatial contexts. A model can learn how to selectively transform features to the latent node by our attentive focus and receive their feedback in the diffusion stage.

\subsubsection{Temporal Context-aware Module}
We exploit the temporal dependencies between learned latent nodes in the graph sequence via Context-Aware Module (CAM) for transforming global spatial-temporal context. Once each spatial graph obtains its $G_S$, the left problem is how to model the temporal contextual information based on these. Long-Short-term Memory (LSTM) \cite{hochreiter1997long} has been validated its strengths in modeling the dependencies and dynamics in sequential data. Hence, we utilize it to build our context-aware module. A LSTM cell contains an input gate $i_t$, a forget gate $f_t$, a output gate $o_t$ and an input cell $g_t$, which are computed by the following functions:
\begin{equation}
\left(
\begin{aligned}
i_t \\
f_t \\
o_t \\
g_t \\
\end{aligned}
\right)
=
\left(
\begin{aligned}
\sigma \\
\sigma \\
\sigma \\
tanh \\
\end{aligned}
\right)
\mathbf{W}
\left(
\begin{aligned}
\hat{\mathbf{G}}^t_S \\
\mathbf{H}_{t-1}
\end{aligned}
\right)
\end{equation}
where $\mathbf{H}_{t-1}$ is the previous hidden state at time $t$, $\mathbf{W} $ is a transformation matrix with learnable parameters, $\sigma$ and $tanh$ are the activation functions. $\hat{\mathbf{G}}^t_S$ is the projected latent node features, which is $\hat{\mathbf{G}}_S = \mathbf{G}_S\mathbf{W}_{2}$ with a weight matrix $\mathbf{W}_{2}\in \mathbb{R}^{C'\times C'\times 1\times 1}$ . The memory cell $c_t$ and hidden state $\mathbf{H}_t$ are updated by :
\begin{equation}
\begin{aligned}
c_t = f_t\otimes c_{t-1} + i_t\otimes g_t \\
\mathbf{H}_t = o_t \otimes tanh(c_t)
\end{aligned}
\end{equation}
where $\otimes$ represents element-wise multiplication operator.

Considering the directed message passing, we stack two bidirectional LSTM cells as our temporal context-aware module. Thus, the learned spatial-temporal context can be formulated as $\mathbf{G}_{ST} =[\overleftarrow{\mathbf{H}}, \overrightarrow{\mathbf{H}]} \in \mathbb{R}^{1\times T\times \hat{C}}$

\subsubsection{Dynamic Attentive Diffusion}
For diffusion stage, we expect the neural network to learn how much spatial-temporal context should be passed back when updating node features. With this diffusion, one node in a single graph can receive the information from inter-frame and intra-frame nodes, which indicates the learned latent node can be regarded as a transformer. 

In details, the net firstly learns a weight matrix $\mathbf{W}_{d} \in \mathbb{R}^{V\times T\times 1}$ for conveying global context to every spatial node in each frame. Then the node $v_i$ absorbs the diffused contextual information $\mathbf{f}_g^{v_i}$ to concatenate with its node embedding $\mathbf{f}_{in}^{v_i}$. This combined node features are embedded via a $1\times 1$ convolution operation with a learnable $\mathbf{W}_3 \in \mathbb{R}^{(C'+\hat{C})\times C'\times1 \times1}$. Hence, the diffusion is formulated as:
\begin{equation}
\begin{aligned}
\mathbf{f}_g^t &= \mathbf{W}_{d}^t \mathbf{G}_{ST}^t \\
\mathbf{f}_{out}&=[\mathbf{f}_{in}, \mathbf{f}_g] \mathbf{W}_3
\end{aligned}
\end{equation}
where $[\cdot , \cdot]$ denotes concatenating along channel dimension.


\section{Experiments}
To evaluate the effectiveness of our approach, we conduct extensive experiments to compare with state-of-the-art methods on two challenging benchmarks: NTU-RGB+D \cite{shahroudy2016ntu} and Skeleton-Kinetics\cite{yan2018spatial}. Considering the limited computing resources, we choose the relative smaller NTU-RGB+D dataset to analyze the effect of components in our Bidirectional Attentive Graph Convolutional Network(BAGCN).
\subsection{Datasets and Setting}
\textbf{NTU-RGB+D:} NTU-RGB+D \cite{shahroudy2016ntu} is collected by Microsoft Kinect v2 with joints annotations. It consists of 56,880 video samples from 60 categories of human actions. 40 distinct subjects perform various actions, which are captured by three cameras simultaneously with different horizontal angles: -$45^o$, $0^o$, $45^o$. The human pose are articulated as 25 joints and their coordinates are labeled as annotations. The evaluation of this dataset is divided into two standard protocols: Cross-Subject (X-sub) and Cross-View (X-view). In the former configuration, half of the 40 subjects consists of the training set and the other for testing. For the latter, training and testing groups are split in terms of the camera views, where the training group has 37920 video samples from camera 2 and 3. Only top-1 accuracy is reported on both of the two protocols. Data preprocessing follows that in \cite{shi2019skeleton, shahroudy2016ntu}.

\textbf{Skeleton-Kinetics:}  DeepMind Kinetics \cite{kay2017kinetics} human action video dataset contains 400 human action classes, with at least 400 video clips for each action taken from the Internet. Its activity categories contains various indoors and outdoors daily actions, like driving car and dying hair. The skeleton-kinetics data was obtained by estimating the location of 18 articulated joints on every frame with public available \emph{OpenPose}\cite{cao2018openpose} toolbox. The annotation of these joints consists of their 2D coordinates and 1D confidence score. We follow the same data preparation strategy in ST-GCN \cite{yan2018spatial} for fair comparison and select two bodies from multi-person cases according to the highest average joints confidence. All the samples are resized to 300 frames by padding. Top-1 and Top-5 classification accuracy are chosen to serve as evaluation metric with 240,000 and 20,000 clips for training and testing, respectively.

\textbf{Implementation details:}
Here, we introduce how to implement our Bidirectional Attentive Graph Convolutional Network (BAGCN) and the training details. We concatenate joints and bones along the channel dimension following \cite{shi2019two} and fed it into networks for almost the experiments.

Our BAGCN consists of 9 building blocks and the first layer have 64 channels for output. In the 4-th and 7-th blocks, we double the number of channels while downsample the temporal length by a factor 2, as the same in ST-GCN\cite{yan2018spatial}. A data BN layer performs the normalization at the beginning and a global average pooling layer follows the last block to generate a 256 dimension  vector for each sequence, which is then classified by a softmax classifier. For the context-aware focusing and diffusion, $C$ and $C^{''}$ is the input and output channels of corresponding blocks, where $C' = \frac{1}{4}C^{''}$ and $\hat{C}=128$. Besides,  $\mathbf{W}_f$ and $\mathbf{W}_d$ share the weight and learn from $\mathbf{f}_{in}$ directly using a $C'\times 1 \times 1 \times 1$ convolution. Stochastic gradient descent with 0.9 momentum, weight decay with 0.0001 and Cross-entropy are applied for training. We initialize the base learning rate as 0.1 and decay it by 10 at $30_{th}$ and $40_{th}$ epoch of 50 epochs for NTU-RGB+D dataset while $20_{th}$, $40_{th}$  $55_{th}$ of 65 epochs for Skeleton-Kinetics dataset. Besides, we also explore the spatial and motion information referred in DGNN \cite{shi2019skeleton}. The motion stream takes the movements of joints and the deformations of bones as input, which is exploited only in the last part in ablation study. Otherwise, all the other experiments are based on the spatial stream for fair comparison.
\subsection{Compared with State-of-the-art Methods}
\begin{table}[t]
	\centering
	\caption{Top-1 accuracy of action recognition on NTU-RGB+D compared with state-of-the-art methods.}
	\begin{tabular}{l|c|c}
		\toprule
		\midrule
		Methods & Cross-Subject & Cross-View \\
		\midrule
		Lie Group \cite{vemulapalli2014human} & 50.1\% & 82.8\% \\
		H-RNN\cite{du2015hierarchical} & 59.1\% & 64.0\% \\
		Deep LSTM\cite{shahroudy2016ntu} & 60.7\% & 67.3\% \\
		ST-LSTM\cite{liu2016spatio} & 69.2\% & 77.7\% \\
		STA-LSTM\cite{song2017end} & 73.4\% &  81.2\%\\
		Temporal Conv\cite{kim2017interpretable} & 74.3\% & 84.1\% \\
		Clip+CNN+MTLN\cite{ke2017new} & 79.6\% & 84.8\% \\
		Synthesized CNN\cite{liu2017enhanced} & 80.0\%& 87.2\% \\
		ST-GCN\cite{yan2018spatial} & 81.5\%& 88.3\%\\
		Two-Stream CNN\cite{li2017skeleton} & 83.2\%& 89.3\%\\
		SR-TSL\cite{si2018skeleton} & 84.8\% & 92.4\% \\
		HCN \cite{li2018co}& 86.5\%& 91.1\%\\ 
		AS-GCN\cite{li2019actional} & 86.8\%& 94.2\% \\
		2s-AGCN\cite{shi2019two} & 88.5\% & 95.1\% \\
		DGNN\cite{shi2019skeleton} & 89.9\%& 96.1\% \\
		\midrule
		BAGCN (Ours) & \textbf{90.3\%} & \textbf{96.3\%} \\
		\midrule
		\bottomrule
	\end{tabular}
	\label{tab:ntu_all}
\end{table}

To validate the superior performance of the BAGCN, we compare it with state-of-the-arts on both NTU-RGB+D and Skeleton-Kinetics datasets. 

For the former, we report the cross-subject and cross-view protocols of all the  compared methods in Tab.\ref{tab:ntu_all}. These approaches can be divided into handcraft-based \cite{vemulapalli2014human}, RNN-based \cite{du2015hierarchical,shahroudy2016ntu, liu2016spatio, song2017end}, CNN-based \cite{kim2017interpretable,li2018co, li2017skeleton,ke2017new, liu2017enhanced} and recent GCN-based\cite{yan2018spatial, si2018skeleton, li2019actional, shi2019two, shi2019skeleton}. Our model outperforms previous methods. Compared with our baseline ST-GCN, the context-aware focusing and diffusion mechanism enhances the performance by a great margin, which increases from 81.5\% to 90.3\% and 88.3\% to 96.3\% in terms of cross-subject and cross-view, respectively. This performance gain verifies the superiority and effectiveness of BAGCN. The effect of components in BAGCN is explored in the later ablation study. 

For Skeleton-Kinetics datasets, we compare with other methods in terms of top-1 and top-5 accuracies in Tab.\ref{tab:kinetics}. We see that our BAGCN achieves the best performance, which is identical to the former. From these two table, it is found that graph-based methods exploit better kinematic dependency than RNN/CNN based ones and our focusing and diffusion mechanism can achieve the message passing between intra-frame joints and inter-frame joints for better recognition accuracies.

\begin{table}[t]
	\centering
	\caption{Comparison with state-of-the-art methods of on Skeleton-Kinetics in terms of Top-1 and Top-5 classification accuracies.}
	\begin{tabular}{c|c|c}
		\toprule
		Methods & Top-1 & Top-5 \\
		\midrule
		Feature Enc. \cite{fernando2015modeling} & 14.9\% & 25.8\% \\
		Deep LSTM \cite{shahroudy2016ntu}& 16.4\% & 35.3\% \\
		Temporal Conv\cite{kim2017interpretable} & 20.3\%& 40.0\%\\
		ST-GCN \cite{yan2018spatial} & 30.7\% & 52.8\& \\
		AS-GCN \cite{li2019actional} & 34.8\% & 56.5\% \\
		2s-AGCN \cite{shi2019two} & 36.1\% & 58.7\% \\
		DGNN \cite{shi2019skeleton} & 36.9\% & 59.6\%\\
		\midrule
		BAGCN(Ours) & \textbf{37.3\%} & \textbf{60.2\%} \\
		\bottomrule
	\end{tabular}
	
	\label{tab:kinetics}
\end{table}

\subsection{Ablation Study}
In this session, we evaluate the effect of components in our bidirectional attentive graph convolutional network. In details, we analyze each component by testing recognition performance of BAGCN without that part on NTU-RGB+D. Besides, we also exploit the impact from different modalities. Compared with the X-view protocol, the X-subject requires a model the ability to recognize actions performed by unseen subjects, which is more challenging. 

\begin{table}[t]
	\caption{Comparison with the baseline ST-GCNs and focusing ways on NTU-RGB+D validated by Cross-Subjects protocol.}
	\setlength{\tabcolsep}{11mm}{
		\begin{tabular}{c|c}
			\toprule
			\midrule
			Methods & Accuracy \\
			\midrule
			\midrule
			ST-GCN$^{\rm *}$  & 81.5\% \\
			ST-GCN  \cite{yan2018spatial} & 84.8\% \\
			\midrule
			A-link GCN$^{\rm +}$ & 83.2\% \\
			S-link GCN$^{\rm +}$ & 84.2\%\\
			AS-GCN \cite{li2019actional}& 86.1\%\\
			\midrule
			BAGCN (wo/F) & 86.4\%\\
			BAGCN (max) & 87.4\% \\
			BAGCN (avg) & 87.8\% \\
			BAGCN (att) & 89.4\%\\
			\midrule
			\bottomrule
		\end{tabular}
	}
	\footnotesize{\\$^{\rm *}$ Original results reported in \cite{yan2018spatial} with only joints information \\ $^{\rm +}$ Actional links (A-link) and Structural links (S-link) learned in \cite{li2019actional}  }
	\label{tab:focus}
\end{table}

\subsubsection{Attentive Focusing and Diffusion}
We analyze the importance of our attentive focusing and diffusion by exploring the different ways of focusing. First, we delete it from our Bidirectional Attentive Graph Convolutional network (BAGCN) and then investigate three types for the focusing stage, results shown in Tab.\ref{tab:focus}.

\textbf{Without attentive focusing and diffusion.}
The original performance in ST-GCN\cite{yan2018spatial} is increased to 84.8\% with the concatenated joints and bones information, serving as our baseline method. According to the graph construction, our BAGCN degrades to bidirectional ST-GCN without the focusing and diffusion, denoted as BAGCN (wo/F). The message first flows out from the center joint to peripheral joints and then flows back after nodes updating. This message passing way improves the performance from 84.8\% to 86.4\%, that indicates the necessity for a joint node receiving information from spatial non-neighbor joints. AS-GCN \cite{li2019actional} also captures intra-frame dependencies beyond physical connection to update node embeddings. It learns actional links (A-link) and structural links (S-link) simultaneously, which causes difficulty in optimizing. The limited performance of AS-GCN (86.1\%) illustrates this problem and it demonstrates the effectiveness and efficiency of our proposed way of message passing in turn. On the other hand, it indicates that dense dependencies make it hard for the net to overcome noise from those irrelevant ones. 

\textbf{With other ways of focusing and diffusion.} For the focus stage, a latent transformer node first passes the spatial contextual information in a frame and then captures temporal dynamics over frames for spatial-temporal contextual information. This global context will be conveyed back to each spatial skeleton graph in the diffusion stage. Two more types of focus way, average and max-pool, are investigated for proving the importance of attentive focus. Recognition performances are shown in the last three lines of Tab.\ref{tab:focus}. 

It is observed that attentive focusing and diffusion can significantly improve performance. For the max-pool configuration, our BAGCN takes embedding features of maximum response node to initialize the latent node in each frame and then captures temporal contextual information between those maximum response nodes. This captured spatial-temporal context is demonstrated useful in augmenting with the spatial node embedding for every frame, owing to the performance increased by \emph{1.0}\% compared with BAGCN (wo/F). However, It is not reasonable to only focus on the maximum responsive node and ignore those physically associated or semantically associated joint nodes in an action. Hence, we can find that the other two configurations bring higher performance improvement. The average focus way, denoted as BAGCN (avg), emphasizes the mean statistics of the joint nodes' response in every graph convolutional layer and finally achieves 87.8\%. Further, BAGCN (att) leans dynamic spatial attention over graph layers for each frame to selectively pass message forth and back between informative joints. It leads to around \emph{6\%} performance gain compared with our baseline ST-GCN. This attentive focusing and diffusion learns a sequence of latent nodes to capture spatial and temporal contextual information from implicit connections and allows spatial nodes learn how much context should be used when updating node embedding.


\begin{figure}[t]
	\begin{center}
		\includegraphics[width=0.3\linewidth]{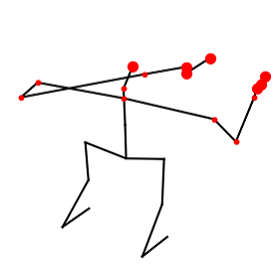}
		\includegraphics[width=0.3\linewidth]{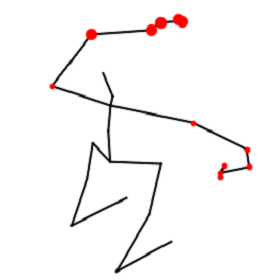}
		\includegraphics[width=0.3\linewidth]{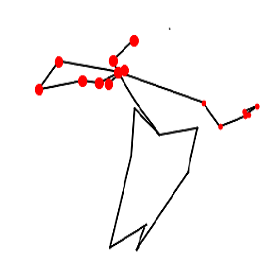}
	\end{center}
	\caption{Illustration of the associated joints with our learned latent transformer node in three action samples. Actions from left to right are reading, hand waving and face wiping, respectively. The red points represent the informative joint nodes and activated joints associated with specific action. The larger circle denotes more informative and relative irrelevant joints are ignored in this figure.}
	\label{fig:latent_node}
\end{figure}
\subsubsection{Latent Edge Visualization}
Various actions my refer to activate different joints while ignore the irrelevant ones. Fig.\ref{fig:latent_node} shows some action samples and their associated joints with our learned latent transformer node. We choose the last graph convolutional layer to display the implicitly sparse dependencies between joint nodes. The attention matrix $\mathbf{W}_f \in \mathbb{R}^{T\times V}$ measures the informative degree of connections between the latent transformer node and joints nodes. In the attentive focusing, if the value of $\mathbf{W}_f^{v_i}$ is larger than 0.8, the connection between latent and joint will be drawn. All the connected joints reflects the activated joint nodes related to a certain action. 

It can be seen that reading action, left part in Fig.\ref{fig:latent_node}, refers to hands and head joints, which agrees with the kinematic dependency. The middle part in Fig.\ref{fig:latent_node} displays the informative joints in hand waving and the right part of Fig.\ref{fig:latent_node} illustrates action-specific related body parts in face wiping.


\subsubsection{Temporal Context-aware or not}

The latent node captures spatial contextual information in a frame and passes its message back after node updating. In this section, we evaluate the effect of temporal context-aware module, which is used to extract the global context in temporal dimension. First, we delete the context-aware and then employ two different types of LSTM for modeling temporal dynamics. The configuration of only learning a latent node with focusing and diffusion graphs obtains the 86.6\% classification accuracy in Tab.\ref{tab:context}. Its performance is only 0.2\% higher than BAGCN (wo/F). In contrast, results in the last two lines demonstrates the power of temporal context-aware module. BAGCN (1-Ca) represents one direction temporal context module achieving 87.3\% and BAGCN (2-Ca) denotes bi-directional one can obtain 89.4\% recognition accuracy, a higher improvement. These improvements validate the effectiveness and necessity of capturing spatial-temporal contextual information for action recognition. 

\begin{table}
	\centering
	\caption{Comparison of classification performance with or without Context-aware (Ca) on NTU-RGB+D validation dataset in Cross-Subjects protocol}
	\setlength{\tabcolsep}{8mm}{
		\begin{tabular}{c|c}
			\toprule
			\midrule
			Methods & Accuracy \\
			\midrule
			ST-GCN \cite{yan2018spatial}  & 84.8\% \\
			BAGCN (wo/F) & 86.4\%\\
			BAGCN (wo/Ca) & 86.6\% \\
			\midrule
			BAGCN (1-Ca) & 87.3\% \\
			BAGCN (2-Ca) & 89.4\%\\
			\midrule
			\bottomrule
		\end{tabular}
	}
	\label{tab:context}
\end{table}

\subsubsection{Two-stream or not}
Inspired by previous works \cite{li2017skeleton,li2018co, shi2019skeleton}, we test the supplementary effect of motion information in our bidirectional attentive graph convolutional network. We perform experiments to train the spatial and motion branch separately and then employ late fusion on NTU-RGB+D dataset in Cross-Subjects and Cross-View protocols, which are shown in Tab.\ref{tab:2s}. From this table, it can be found that the spatial branch achieves higher classification performance than the motion one while fusing them together can still improve the accuracy from 89.4\% to 90.3\% in Cross-subject and 95.6\% to 96.3\% in cross-view.

\begin{table}
	\centering
	\caption{Comparison of classification performance with spatial information, motion information and their fusion on NTU-RGB+D validation dataset in Cross-Subjects and Cross-views protocols and Skeleton-Kinetics in top-1,5 accuracies}
	\setlength{\tabcolsep}{3mm}{
		\begin{tabular}{c|c|c|c|c}
			\toprule
			\midrule
			Methods & X-Sub & X-View & Top-1 & Top5\\
			\midrule
			\midrule
			Spatial & 89.4\% & 95.6\% & 36.4\% & 58.9\%\\
			Motion & 86.3\% & 94.1\% & 31.8\% & 54.9\%\\
			Fusion & 90.3\% & 96.3\% & 37.3\%& 60.2\%\\
			\midrule
			\bottomrule
		\end{tabular}
	}
	\label{tab:2s}
\end{table}

\section{Conclusion}
In this paper, we introduce a focusing and diffusion mechanism to enhance the graph convolutional network for receiving the information beyond the spatial and temporal neighbors restrictions. Besides, we propose a context-aware module to capture the global spatial-temporal contextual information for modeling implicit correlations between inter-frame and intra-frame joints. We then develop a Bidirectional Attentive Graph Convolutional Network (BAGCN) as an exemplar. This graph-based model is enhanced with our context-aware focusing and diffusion mechanism. Its superior performance is demonstrated on two public challenging skeleton-based action recognition benchmarks: NTU-RGB+D and Skeleton-Kinetics.

{\small
\bibliographystyle{ieee_fullname}
\bibliography{main}

\begin{thebibliography}{10}\itemsep=-1pt

\bibitem{bruna2013spectral}
Joan Bruna, Wojciech Zaremba, Arthur Szlam, and Yann LeCun.
\newblock Spectral networks and locally connected networks on graphs.
\newblock {\em arXiv preprint arXiv:1312.6203}, 2013.

\bibitem{cao2018openpose}
Zhe Cao, Gines Hidalgo, Tomas Simon, Shih-En Wei, and Yaser Sheikh.
\newblock Openpose: realtime multi-person 2d pose estimation using part
  affinity fields.
\newblock {\em arXiv preprint arXiv:1812.08008}, 2018.

\bibitem{du2015hierarchical}
Yong Du, Wei Wang, and Liang Wang.
\newblock Hierarchical recurrent neural network for skeleton based action
  recognition.
\newblock In {\em Proceedings of the IEEE conference on computer vision and
  pattern recognition}, pages 1110--1118, 2015.

\bibitem{fernando2015modeling}
Basura Fernando, Efstratios Gavves, Jose~M Oramas, Amir Ghodrati, and Tinne
  Tuytelaars.
\newblock Modeling video evolution for action recognition.
\newblock In {\em Proceedings of the IEEE Conference on Computer Vision and
  Pattern Recognition}, pages 5378--5387, 2015.

\bibitem{he2019geonet}
Tong He, Haibin Huang, Li Yi, Yuqian Zhou, Chihao Wu, Jue Wang, and Stefano
  Soatto.
\newblock Geonet: Deep geodesic networks for point cloud analysis.
\newblock In {\em Proceedings of the IEEE Conference on Computer Vision and
  Pattern Recognition}, pages 6888--6897, 2019.

\bibitem{He2019Mono3DM3}
Tong He and Stefano Soatto.
\newblock Mono3d++: Monocular 3d vehicle detection with two-scale 3d hypotheses
  and task priors.
\newblock In {\em AAAI}, 2019.

\bibitem{henaff2015deep}
Mikael Henaff, Joan Bruna, and Yann LeCun.
\newblock Deep convolutional networks on graph-structured data.
\newblock {\em arXiv preprint arXiv:1506.05163}, 2015.

\bibitem{hochreiter1997long}
Sepp Hochreiter and J{\"u}rgen Schmidhuber.
\newblock Long short-term memory.
\newblock {\em Neural computation}, 9(8):1735--1780, 1997.

\bibitem{hussein2013human}
Mohamed~E Hussein, Marwan Torki, Mohammad~A Gowayyed, and Motaz El-Saban.
\newblock Human action recognition using a temporal hierarchy of covariance
  descriptors on 3d joint locations.
\newblock In {\em Twenty-Third International Joint Conference on Artificial
  Intelligence}, 2013.

\bibitem{jiang2015human}
Yu-Gang Jiang, Qi Dai, Wei Liu, Xiangyang Xue, and Chong-Wah Ngo.
\newblock Human action recognition in unconstrained videos by explicit motion
  modeling.
\newblock {\em IEEE Transactions on Image Processing}, 24(11):3781--3795, 2015.

\bibitem{kay2017kinetics}
Will Kay, Joao Carreira, Karen Simonyan, Brian Zhang, Chloe Hillier, Sudheendra
  Vijayanarasimhan, Fabio Viola, Tim Green, Trevor Back, Paul Natsev, et~al.
\newblock The kinetics human action video dataset.
\newblock {\em arXiv preprint arXiv:1705.06950}, 2017.

\bibitem{ke2017new}
Qiuhong Ke, Mohammed Bennamoun, Senjian An, Ferdous Sohel, and Farid Boussaid.
\newblock A new representation of skeleton sequences for 3d action recognition.
\newblock In {\em Proceedings of the IEEE conference on computer vision and
  pattern recognition}, pages 3288--3297, 2017.

\bibitem{kim2017interpretable}
Tae~Soo Kim and Austin Reiter.
\newblock Interpretable 3d human action analysis with temporal convolutional
  networks.
\newblock In {\em 2017 IEEE conference on computer vision and pattern
  recognition workshops (CVPRW)}, pages 1623--1631. IEEE, 2017.

\bibitem{kipf2018neural}
Thomas Kipf, Ethan Fetaya, Kuan-Chieh Wang, Max Welling, and Richard Zemel.
\newblock Neural relational inference for interacting systems.
\newblock {\em arXiv preprint arXiv:1802.04687}, 2018.

\bibitem{kipf2016semi}
Thomas~N Kipf and Max Welling.
\newblock Semi-supervised classification with graph convolutional networks.
\newblock {\em arXiv preprint arXiv:1609.02907}, 2016.

\bibitem{li2017skeleton}
Chao Li, Qiaoyong Zhong, Di Xie, and Shiliang Pu.
\newblock Skeleton-based action recognition with convolutional neural networks.
\newblock In {\em 2017 IEEE International Conference on Multimedia \& Expo
  Workshops (ICMEW)}, pages 597--600. IEEE, 2017.

\bibitem{li2018co}
Chao Li, Qiaoyong Zhong, Di Xie, and Shiliang Pu.
\newblock Co-occurrence feature learning from skeleton data for action
  recognition and detection with hierarchical aggregation.
\newblock {\em arXiv preprint arXiv:1804.06055}, 2018.

\bibitem{li2019actional}
Maosen Li, Siheng Chen, Xu Chen, Ya Zhang, Yanfeng Wang, and Qi Tian.
\newblock Actional-structural graph convolutional networks for skeleton-based
  action recognition.
\newblock In {\em Proceedings of the IEEE Conference on Computer Vision and
  Pattern Recognition}, pages 3595--3603, 2019.

\bibitem{li2015gated}
Yujia Li, Daniel Tarlow, Marc Brockschmidt, and Richard Zemel.
\newblock Gated graph sequence neural networks.
\newblock {\em arXiv preprint arXiv:1511.05493}, 2015.

\bibitem{liu2015simple}
Fang Liu, Xiangmin Xu, Shuoyang Qiu, Chunmei Qing, and Dacheng Tao.
\newblock Simple to complex transfer learning for action recognition.
\newblock {\em IEEE Transactions on Image Processing}, 25(2):949--960, 2015.

\bibitem{liu2016spatio}
Jun Liu, Amir Shahroudy, Dong Xu, and Gang Wang.
\newblock Spatio-temporal lstm with trust gates for 3d human action
  recognition.
\newblock In {\em European Conference on Computer Vision}, pages 816--833.
  Springer, 2016.

\bibitem{liu2017enhanced}
Mengyuan Liu, Hong Liu, and Chen Chen.
\newblock Enhanced skeleton visualization for view invariant human action
  recognition.
\newblock {\em Pattern Recognition}, 68:346--362, 2017.

\bibitem{liu2019flownet3d}
Xingyu Liu, , Charles~R Qi, and Leonidas~J Guibas.
\newblock Flownet3d: Learning scene flow in 3d point clouds.
\newblock {\em Proc. Computer Vision and Pattern Recognition (CVPR), IEEE},
  2019.

\bibitem{mehran2009abnormal}
Ramin Mehran, Alexis Oyama, and Mubarak Shah.
\newblock Abnormal crowd behavior detection using social force model.
\newblock In {\em 2009 IEEE Conference on Computer Vision and Pattern
  Recognition}, pages 935--942. IEEE, 2009.

\bibitem{niepert2016learning}
Mathias Niepert, Mohamed Ahmed, and Konstantin Kutzkov.
\newblock Learning convolutional neural networks for graphs.
\newblock In {\em International conference on machine learning}, pages
  2014--2023, 2016.

\bibitem{qi2017pointnet}
Charles~R Qi, Hao Su, Kaichun Mo, and Leonidas~J Guibas.
\newblock Pointnet: Deep learning on point sets for 3d classification and
  segmentation.
\newblock In {\em Proceedings of the IEEE Conference on Computer Vision and
  Pattern Recognition}, pages 652--660, 2017.

\bibitem{qi2017pointnet++}
Charles~Ruizhongtai Qi, Li Yi, Hao Su, and Leonidas~J Guibas.
\newblock Pointnet++: Deep hierarchical feature learning on point sets in a
  metric space.
\newblock In {\em Advances in neural information processing systems}, pages
  5099--5108, 2017.

\bibitem{shahroudy2016ntu}
Amir Shahroudy, Jun Liu, Tian-Tsong Ng, and Gang Wang.
\newblock Ntu rgb+ d: A large scale dataset for 3d human activity analysis.
\newblock In {\em Proceedings of the IEEE conference on computer vision and
  pattern recognition}, pages 1010--1019, 2016.

\bibitem{shi2019skeleton}
Lei Shi, Yifan Zhang, Jian Cheng, and Hanqing Lu.
\newblock Skeleton-based action recognition with directed graph neural
  networks.
\newblock In {\em Proceedings of the IEEE Conference on Computer Vision and
  Pattern Recognition}, pages 7912--7921, 2019.

\bibitem{shi2019two}
Lei Shi, Yifan Zhang, Jian Cheng, and Hanqing Lu.
\newblock Two-stream adaptive graph convolutional networks for skeleton-based
  action recognition.
\newblock In {\em Proceedings of the IEEE Conference on Computer Vision and
  Pattern Recognition}, pages 12026--12035, 2019.

\bibitem{si2018skeleton}
Chenyang Si, Ya Jing, Wei Wang, Liang Wang, and Tieniu Tan.
\newblock Skeleton-based action recognition with spatial reasoning and temporal
  stack learning.
\newblock In {\em Proceedings of the European Conference on Computer Vision
  (ECCV)}, pages 103--118, 2018.

\bibitem{song2017end}
Sijie Song, Cuiling Lan, Junliang Xing, Wenjun Zeng, and Jiaying Liu.
\newblock An end-to-end spatio-temporal attention model for human action
  recognition from skeleton data.
\newblock In {\em Thirty-first AAAI conference on artificial intelligence},
  2017.

\bibitem{soomro2012ucf101}
Khurram Soomro, Amir~Roshan Zamir, and Mubarak Shah.
\newblock Ucf101: A dataset of 101 human actions classes from videos in the
  wild.
\newblock {\em arXiv preprint arXiv:1212.0402}, 2012.

\bibitem{torresani2004learning}
Lorenzo Torresani, Aaron Hertzmann, and Christoph Bregler.
\newblock Learning non-rigid 3d shape from 2d motion.
\newblock In {\em Advances in Neural Information Processing Systems}, pages
  1555--1562, 2004.

\bibitem{vemulapalli2014human}
Raviteja Vemulapalli, Felipe Arrate, and Rama Chellappa.
\newblock Human action recognition by representing 3d skeletons as points in a
  lie group.
\newblock In {\em Proceedings of the IEEE conference on computer vision and
  pattern recognition}, pages 588--595, 2014.

\bibitem{wang2012mining}
Jiang Wang, Zicheng Liu, Ying Wu, and Junsong Yuan.
\newblock Mining actionlet ensemble for action recognition with depth cameras.
\newblock In {\em 2012 IEEE Conference on Computer Vision and Pattern
  Recognition}, pages 1290--1297. IEEE, 2012.

\bibitem{wang2016temporal}
Limin Wang, Yuanjun Xiong, Zhe Wang, Yu Qiao, Dahua Lin, Xiaoou Tang, and Luc
  Van~Gool.
\newblock Temporal segment networks: Towards good practices for deep action
  recognition.
\newblock In {\em European conference on computer vision}, pages 20--36.
  Springer, 2016.

\bibitem{wu2019comprehensive}
Zonghan Wu, Shirui Pan, Fengwen Chen, Guodong Long, Chengqi Zhang, and Philip~S
  Yu.
\newblock A comprehensive survey on graph neural networks.
\newblock {\em arXiv preprint arXiv:1901.00596}, 2019.

\bibitem{yan2018spatial}
Sijie Yan, Yuanjun Xiong, and Dahua Lin.
\newblock Spatial temporal graph convolutional networks for skeleton-based
  action recognition.
\newblock In {\em Thirty-Second AAAI Conference on Artificial Intelligence},
  2018.

\bibitem{yun2012two}
Kiwon Yun, Jean Honorio, Debaleena Chattopadhyay, Tamara~L Berg, and Dimitris
  Samaras.
\newblock Two-person interaction detection using body-pose features and
  multiple instance learning.
\newblock In {\em 2012 IEEE Computer Society Conference on Computer Vision and
  Pattern Recognition Workshops}, pages 28--35. IEEE, 2012.

\end{thebibliography}
}

\end{document}